\documentclass[11pt,a4paper]{article}
\usepackage{times,latexsym}
\usepackage{amsmath}
\usepackage{url}
\usepackage[T1]{fontenc}
\usepackage{graphicx}
\usepackage{algorithm}
\usepackage{algpseudocode}
\usepackage{listings}
\usepackage{wrapfig}
\usepackage{verbatim}
\usepackage{textcomp}
\usepackage{makecell}
\usepackage{float}
\usepackage{color}
\usepackage{soul}
\usepackage[usenames,dvipsnames]{xcolor}
\usepackage{bm}
\usepackage{adjustbox}
\usepackage{array}
\usepackage{float}

\newcolumntype{R}[2]{%
    >{\adjustbox{angle=#1,lap=\width-(#2)}\bgroup}%
    l%
    <{\egroup}%
}

\setcellgapes{4pt}

\usepackage[acceptedWithA]{tacl2021v1}

\usepackage{xspace,mfirstuc,tabulary}
\usepackage{multirow}

\newif\iftaclinstructions
\taclinstructionsfalse 
\iftaclinstructions
\renewcommand{\confidential}{}
\renewcommand{\anonsubtext}{(No author info supplied here, for consistency with
TACL-submission anonymization requirements)}
\newcommand{\instr}
\fi

\usepackage{paralist}

\newcommand\blfootnote[1]{%
  \begingroup
  \renewcommand\thefootnote{}\footnote{#1}%
  \addtocounter{footnote}{-1}%
  \endgroup
}

\iftaclpubformat 

\else

\fi

\title{Addressing Topic Leakage in Cross-Topic Evaluation \\ for Authorship Verification}

\author{
  Jitkapat Sawatphol \hspace{1em} \
  Can Udomcharoenchaikit\textsuperscript{ *} \hspace{1em} \ 
   Sarana Nutanong\textsuperscript{ *} 
  \ \\
  School of Information Science and Technology
  \\
  Vidyasirimedhi Institute of Science and Technology, Thailand
  \\
  \texttt{\{jitkapat.s\_s20, canu\_pro, snutanon\}@vistec.ac.th}
  \\
}

\date{}

\begin{document}
\maketitle
\blfootnote{\textsuperscript{*}Corresponding Authors}
\begin{abstract}
Authorship verification (AV) aims to identify whether a pair of texts has the same author. We address the challenge of evaluating AV models' robustness against topic shifts. The conventional evaluation assumes minimal topic overlap between training and test data. However, we argue that there can still be topic leakage in test data, causing misleading model performance and unstable rankings. To address this, we propose an evaluation method called Heterogeneity-Informed Topic Sampling (HITS), which creates a smaller dataset with a heterogeneously distributed topic set. Our experimental results demonstrate that HITS-sampled datasets yield a more stable ranking of models across random seeds and evaluation splits. Our contributions include: 1. An analysis of causes and effects of topic leakage. 2. A demonstration of the HITS in reducing the effects of topic leakage, and 3. The Robust Authorship Verification bENchmark (RAVEN) that allows topic shortcut test to uncover AV models' reliance on topic-specific features.
\end{abstract}

\section{Introduction}

Authorship verification (AV) is a task that aims to predict whether a pair of texts is written by the same author. 
A common research problem in AV is to develop a model that performs well across unseen topics, domains, or genres \cite{mikros2007investigating, stamatatos2013robustness, sapkota2014cross}. Our study focuses on the unseen topic problem for two reasons: First, it is not realistic to assume that an author will always write on the same topics. Second, the unseen texts might be written on topics with unseen keywords or themes. Unlike typical domain adaptation scenarios, a topic shift can also be subtle and unrecognized. We think it is useful for AV systems to be able to recognize authors’ writing styles regardless of whether the topics of texts change or not.  

To develop such systems, cross-topic benchmarks are necessary to assess the model’s performance in handling topic shifts and compare the effectiveness of various methods. Existing cross-topic AV evaluations assume that different topic categories hold dissimilar information. 
Consequently, the topic shift is commonly simulated by separating training and test data across two different sets of topics. 
For example, two topics are automatically considered dissimilar if they come from two different domain categories.  
%
%

%
In this paper, we challenge the conventional practice of cross-topic split by viewing topic similarity as a value on a continuous spectrum.
In other words, some pairs of topics may be \emph{more similar}, sharing common attributes or characteristics than the rest.  
We argue that performing a cross-topic train-test split without considering topic similarity can cause topic leakage. 
Topic leakage has been suggested \cite{sawatphol-etal-2022-topic} to exist when documents in cross-topic test data unintentionally contain topical information similar to those in training data. Furthermore, we explain that topic leakage can cause uncertainties in model evaluation and selection. For example, some models may rely on only learning shortcuts from topic-specific features. Such models can demonstrate inflated performances in test data with topic leakage.

To address the issue of topic leakage, we propose an evaluation framework that takes into account the similarity of topics in a dataset. The crux of our method lies in the similarity-based sampling technique that can create a smaller but more topically heterogeneous version of any existing dataset. This ensures that each topic category is less overlapping in information, thus helping reduce topic leakage. Our experimental results demonstrate that our evaluation method can help prevent misleading cross-topic performance by exposing models relying on topic shortcuts and improving model ranking stability compared to comparable-sized datasets with randomly distributed topics. 

We summarize our contribution as follows.
\begin{enumerate}
    \item We introduce the notion of topic similarity, which can cause information leakage between train and test data in creating cross-topic benchmarks for the AV task. 
    \item We propose \emph{Heterogeneity-Informed Topic Sampling (HITS)}, a framework that considers the topical heterogeneity in the dataset to mitigate possible topic leakage issues. This framework can be applied to any existing dataset to improve topic shift degree and ranking stability compared to conventional cross-topic evaluation approaches.
    \item We provide \emph{RAVEN}, a benchmark comprising datasets with heterogeneous topic sets. This benchmark can be beneficial toward the development of topic-robust AV methods by allowing the identification and comparison of models' reliance on topic-specific shortcuts.
\end{enumerate}

\section{Related Works}
A limited number of works have attempted to provide a standard benchmark to compare the effectiveness of AV methods on cross-topic setups.  In particular, the PAN 2020 and 2021 AV task \cite{Kestemont2020OverviewOT, kestemont2021overview} is considered one of the largest benchmarks for cross-topic AV that compares the effectiveness of many AV approaches. The PAN organizers use a dataset collected from fanfiction.com comprising over 4,000 topics. Additionally, \citet{brad-etal-2022-rethinking} introduced a cross-topic setup (denoted as \emph{open unseen fandoms}) as one of their evaluation splits that extend the experiment setups of PAN competitions. \citet{tyo2022state} also attempted to compare various attribution and verification methods on the 15 datasets, but only four were cross-topic evaluations. In more extreme cases, the study by \citet{altakrori-etal-2021-topic-confusion} tests authorship attribution models' behavior by deliberately inverting the topic-author relationship between training and test data, a setup considered more challenging than regular cross-topic train-test splits. 
 
Other than benchmarks, numerous studies have also attempted to develop AV systems to handle topical changes in texts. Those studies conducted experiments with many different variations on datasets and problem formulations. For example, \citet{mikros2007investigating} has studied the features' topic independence using Greek newspaper articles. In addition, \citet{stamatatos2017authorship} has proposed using text distortion to mask topic-specific terms, having experimented with datasets collected from \textit{the Guardian} news articles. More recently, \citet{boenninghoff2021o2d2} has proposed a hybrid neural-probabilistic system, achieving state-of-the-art performance on the PAN2021's cross-topic AV competition. Furthermore, \citet{rivera-soto-etal-2021-learning} and \citet{sawatphol-etal-2022-topic} studied applying representation learning in authorship verification, evaluating in test sets with both unseen topics and unseen authors. 

Despite advances in cross-topic AV research, few studies have questioned the limitations of existing cross-topic evaluation. The most closely related study we have found is conducted by \citet{wegmann-etal-2022-author}. The study suggests that current AV training objectives either do not control content or only use domain/topic labels to approximate topic differences. The study then proposed a contrastive style representation learning task that controls the topic similarity of each text pair to help force models to favor learning writing styles rather than content. In our understanding, content control is a broader concept similar to cross-topic evaluation. However, the key difference is that their work primarily addresses content control between text pairs to improve style representation learning of AV models. On the other hand, our study aims to control the topic information at the dataset level to enhance the reliability of cross-topic evaluation.

\begin{figure*}[t]
\centering
\includegraphics[width=\textwidth]{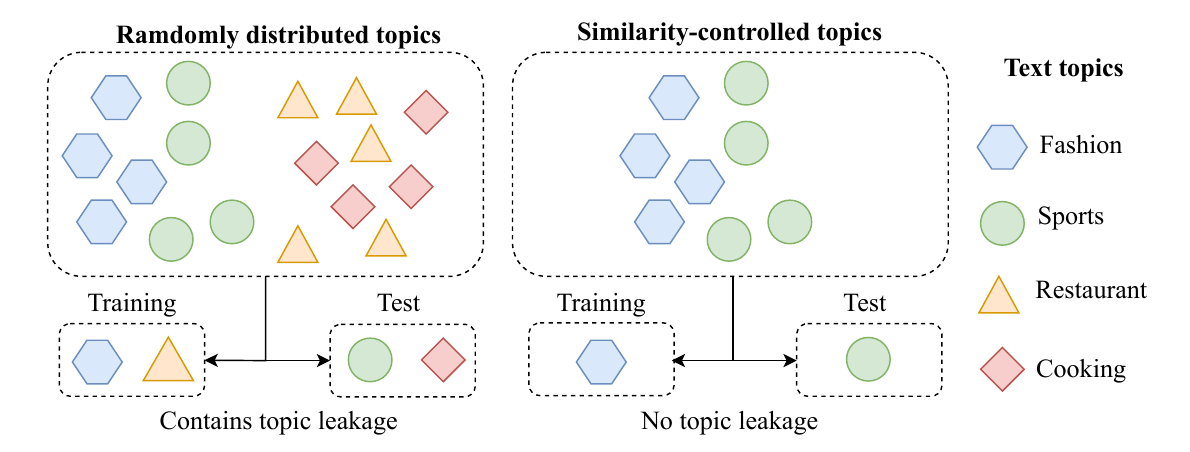}
\caption{An illustration of two different scenarios of how texts in various topics are distributed on a topic embedding space. In randomly distributed datasets, some topics are more similar to each other than other topics. When performing a cross-topic train-test split from this data, some topics are leaked into the test data. On the other hand, a topic similarity-controlled dataset removes topics that are similar to each other, reducing the degree of leakage}.

\label{heterogeneityfig}
\end{figure*}

To the best of our knowledge, in studies proposing either cross-topic benchmarks or cross-topic AV methods, we notice that these studies are still limited to the assumption that each labeled topic category is mutually exclusive. Our study argues that such an assumption might lead to an overlooked issue of topic information leakage and its consequences toward model evaluation and selection, which we will further describe in Section \ref{hetero}.

\section{Topic Leakage}
\label{hetero}

In cross-topic AV studies, it is essential to simulate an environment comprising texts from unseen topics to assess models' behavior when applied to topic-shifted scenarios. However, we argue that there is an issue that might diminish the effectiveness of cross-topic evaluation: \emph{topic leakage}.

We define topic leakage as a phenomenon when some topics in test data unintentionally share information with topics in training data. A topic in training data and another in test data might share common topical attributes despite being labeled as different topics. Consequently, the test data may include texts intended to represent unseen topics but are not ``unseen'' regarding topic content. With topic leakage, the ``cross-topic'' property of test sets is diminished.

\paragraph{Causes of Leakage}
Topic leakage is caused by the assumption of topic heterogeneity in AV datasets. These datasets often contain metadata that categorizes texts into specific topics. Researchers leverage these datasets for evaluating models' robustness by implementing a train-test split, where the test data includes texts in topics not in the training set. This conventional evaluation approach assesses models' capacity to work with texts without relying on topic-specific features present in the training data. However, the approach presupposes that texts in each topic category are mutually exclusive, which is not always the case. Figure \ref{heterogeneityfig} illustrates how a collection of randomly distributed topics can cause topic leakage after train-test splits, while a collection of heterogeneous topics can help prevent topic leakage. In this example, the restaurant topic is in training data, and the cooking topic is in test data, despite these two topics having similar content.  This overlap diminishes the intended distribution shift in cross-topic evaluation, as some test data are still similar to topics in the training data. Furthermore, recent evidence, as reported by \citet{sawatphol-etal-2022-topic}, suggests topic information leakage in the train-test split of the Fanfiction dataset in PAN2021 AV competition \cite{kestemont2021overview} where training and test data contain examples of topics sharing information like entity mentions and keywords. As a result, their experimental results show similar cross-topic evaluation performance to the in-distribution-topic experiments.

\begin{figure*}[t]
\centering
\includegraphics[width=\textwidth]{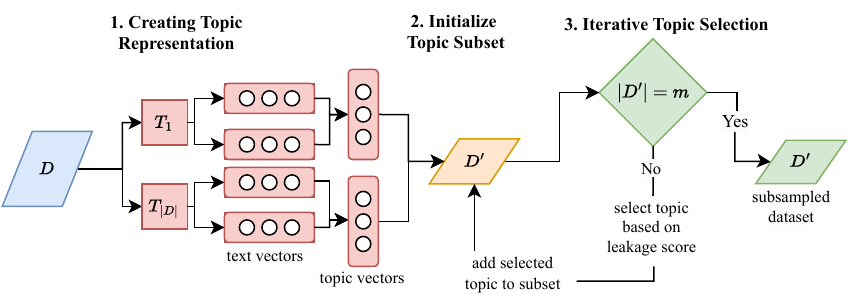}
\caption{A diagram illustrating the pipeline of the HITS method in selecting a topic-heterogeneous subset.}
\label{hitspipeline}
\end{figure*}

\paragraph{Consequences.} We argue that topic leakage can lead to the following negative consequences: misleading evaluation and unstable model rankings. To the best of our knowledge, these issues are not commonly discussed in existing studies. 

\emph{Misleading evaluation.} Topic leakage can complicate measuring a model's performance in topic-shifted scenarios. A model may show strong performance on a ``cross-topic'' benchmark, implying that it is robust against the topic shift. However, the model might rely on topic bias or spurious correlations between topic-specific keywords and authors rather than learning to distinguish writing styles. This scenario contradicts the objective of cross-topic AV evaluation, which is to build an AV system that works in texts with unseen topics. When there is a risk of topic leakage, the evaluation results can be misled by spurious correlations, which misrepresents the models' robustness against topic shifts in real-world applications.
    
\emph{Unstable model rankings.} Topic leakage can also affect the selection of the most suitable model among candidates. When topic leakage is present in a cross-topic evaluation, a model might erroneously appear to perform better due to spurious correlations. The same model may fail to perform adequately in cross-topic data without leakage. This inconsistency in model performance complicates the model selection process and introduces uncertainty. If a set of candidate models is evaluated on a topic-leaked split, the best-performing model might not be the most robust.

With the issue of possible topic leakage and heterogeneity assumption in mind, our objective is to mitigate the topic information leakage problem. To achieve such goals, we design a method to help ensure the heterogeneous topic categories in datasets, which will be described in Section \ref{proposedmethod}. 
\section{Proposed Evaluation Method}
\label{proposedmethod}

We hypothesize that a more controlled, topic-heterogeneous dataset is less prone to topic information leakage, regardless of the train-test split method. 
The reason behind this hypothesis is that if there is less overlap in information between each topic category, there would be a higher degree of distribution shift in cross-topic evaluation splits.

We present \emph{Heterogeneity Informed Topic Sampling (HITS)}, an evaluation framework involving a subsampling technique that ensures that the resulting subsampled dataset has less topic similarity and, thus, less topic leakage.  
Our framework aims to process a full, original dataset (denoted as $D$) into a smaller but more topic-heterogeneous subset, $D'$. 
To ensure that the resulting subset has low topic similarity, we use an iterative process that selects each candidate topic based on its similarity with previously selected topics. 
The pipeline of the HITS method is illustrated in Figure \ref{hitspipeline}. 

\paragraph{1. Creating topic representation.}  
First, let us denote the dataset as $D$, a set comprising $|D|$ topics $\{T_1, ..., T_{|D|}\}$. 
Each topic $T_{i}$ is a set containing $|T_{i}|$ vectors representing each text document in that topic. 
The vectors can be created with any encoding function. We use a pre-trained SentenceBERT \cite{reimers-gurevych-2019-sentence} model in our experimental studies in this paper. 
We use $\bm{v}_{i,k}$ to denote the representation of text $k$ within topic $i$. 
To create a vector representing each topic, $\bm{t}_{i}$, we compute the mean of the vectors of all texts within that topic as shown in Equation~\ref{gettopicvec}. 

\begin{equation}
{\bm{t}_{i}} = \frac{1}{|T_{i}|} \sum^{|T_{i}|}_{k=1}\bm{v}_{i,k}
\label{gettopicvec}
\end{equation}

\paragraph{2. Initialize topic subset.} 
Since we want to select topics based on their similarity to the previously selected topics, we need a separate set with an initial topic. First, we initialize $D'$ as an empty set. We then compute the mean average cosine similarity between each topic and every other topic in the dataset. Afterward, we select the topic with the lowest mean average cosine similarity with other topics in $D$ and add it to $D'$.

\paragraph{3. Iterative Topic Selection.} We then iteratively select a topic to add to $D'$. The following steps are repeated until the number of topics in $D'$ reaches $m$, where $m$ is a manually set parameter.

\begin{enumerate}
    \item First, we compute the cosine similarity between each topic in $D$ and each topic in $D'$. We aim to select a topic that is the least similar to the previously selected topics. We denote $S_{i}$ as a set of similarities of a topic $i$ each of the previously selected topic $j$.

    \begin{equation}
    S_{i} = \{\text{sim}(\bm{t}_{i}, \bm{t}_{j}), j = 1, ..., |D'|\}
    \label{computingtopicsim}
    \end{equation}

    \item Second, for each topic $T_i$, we compute the leakage score of that topic as described in Equation \ref{leakagescore}. We denote $l_i$ as the leakage score of the topic $T_i$. 

    \begin{equation}
    l_{i} = \text{mean}(S_{i}) * \text{max}(S_{i})
    \label{leakagescore}
    \end{equation}
    
    To prevent a possible scenario of two topics having high similarity but low similarity with the rest, we compute the leakage score with the mean similarity scaled by max similarity.

    The intuition behind this score is that if the representation of that topic is similar to other topics, that topic is closely related to other topics and is more likely to cause leakage. 

    \item Lastly, we select one topic with the lowest leakage score (that is \emph{not} already present in $D'$) and add that topic to $D'.$ the leakage score will be recomputed for the next iteration since the members of $D'$ are updated.
\end{enumerate}

After $m-1$ iterations, the size of $D'$ reaches $m$. We discard other topics from the original dataset $D$. 
We also considered merging unselected topics with these selected topics, but experiments in Section \ref{ablationsec} suggest that discarding the unselected topics yields better ranking stability.

Finally, we obtain $D'$, a topic-heterogeneous version of the original dataset $D$ with reduced topic information leakage. 
We can use the HITS method to convert any existing dataset on any task with topic or domain category labels into an evaluation dataset with less topic information leakage.

\section{Experimental Setup}
\label{expsetup}

We conducted a number of experimental studies to reveal the consequences of topic leakage described in Section \ref{hetero}: misleading model evaluation and unstable model rankings. 

\subsection{Dataset}
In this study, we use the Fanfiction dataset from the PAN2020 \cite{Kestemont2020OverviewOT} and 2021 competitions \cite{kestemont2021overview}. This dataset comprises fiction texts written by online users on fanfiction.net. The topic category in this dataset is called ``fandom'', which is the source story on which each fiction text is based. The original dataset contains approximately 4,000 fandoms and 50,000 authors. Given the original dataset, we study the difference in creating evaluation splits from subsampled datasets under two conditions:

\begin{enumerate}
    \item \textbf{Similarity-Controlled Topics.}  This condition simulates a dataset with heterogeneous topic categories, revealing whether the misleading assessment and inconsistent rankings are reduced when topic similarity is controlled. To create this condition, we sample a dataset into a topic-heterogeneous using the HITS method from Section \ref{proposedmethod}.

    \item \textbf{Random Topics.} This condition simulates the same distribution of topic information with the original dataset. We do not use the full original dataset since we cannot control the number of documents and authors, all of which might affect the results. The randomly subsampled datasets are used to make results comparable with the HITS version. 
\end{enumerate}

To study the effect of topic heterogeneity in different numbers of topics, we create the sub-datasets with the number of documents and topics as shown in Table \ref{ffspstats}.

\begin{table}[t]
\centering
\resizebox{\columnwidth}{!}{%
\begin{tabular}{@{}c|rrrrrrr@{}}

\multirow{8}{*}{HITS} &
  \multicolumn{1}{r}{$m$} &
  \multicolumn{1}{r}{train pairs} &
  \multicolumn{1}{r}{test pairs} &
  \multicolumn{1}{r}{train auths} &
  \multicolumn{1}{r}{test auths} \\ \hline 
                                 & 50  & 29646 & 3211 & 18046 & 2081 \\
                                 & 60  & 34902 & 3762	 & 22042 & 2555 \\ 
                                 & 70  & 41772 & 4480 & 24956 & 2920 \\ 
                                 & 80  & 47417	 & 5100 & 28668 & 3347 \\ 
                                 & 90 & 53617 & 5734 & 32007 & 3772 & \\ 
                                 & 100 & 60006 & 	6437 &	35197 &	4135	\\ \hline 
\multirow{6}{*}{Random} 
                                 & 50  & 29420 & 3210 & 18205 & 2082 \\ 
                                 & 60  & 35245 & 3840 & 21804 & 2500 \\ 
                                 & 70  & 41177 & 4449 & 25347 & 2933 \\ 
                                 & 80  & 47279 & 5107 & 21804 & 3334 \\ 
                                 & 90 & 53649 & 5784 & 31944 & 3727 \\
                                 & 100 & 59907 & 6443 &	35251 &	4125 \\	
\end{tabular}%
}
\caption{Dataset statistics of our HITS and randomly subsampled Fanfiction datasets, using the number of topics at [50, 60, 70, 80, 90, 100]. Each figure is a rounded mean from 10-fold evaluation splits from each subsampled dataset.  $m$ denotes the number of topics. ``pairs'' denote the number of text pairs in the training and test data. ``auths'' denote the number of authors.} 
\label{ffspstats}
\end{table}

\paragraph{Preprocessing.} We use the following steps. First, we subsample the original dataset by selecting a number of topic subset (parameter $m$) using either HITS or random subsampling. In Section \ref{expresults}, we primarily present the results from datasets with 70 topics. However, in Section \ref{ablationsec}, we also experimented with using 50, 60, 80, 90, and 100 topics to see the effect of different numbers of topics. The dataset statistics are reported in Table \ref{ffspstats}. Second, to create different evaluation splits, we divide the data using the k-fold validation split method with $k$=10. Each validation fold comprises a different set of $k$ topics, and one fold is used for evaluation while the rest is used for training.

\begin{table*}[t]
\centering
\begin{tabular}{c|l|llll|l}
\hline
 Subsampling &
 Method &
   
  AUC &
  c@1 &
  F0.5u &
  F1 &
  Overall \\ \hline
\multirow{5}{*}{HITS} \rule{0pt}{2.5ex} & 
       CharNGram &
  0.964\textsuperscript{\textpm 0.011} &
  \textbf{0.959}\textsuperscript{\textpm 0.011}&
  \textbf{0.859}\textsuperscript{\textpm 0.031}&
  \textbf{0.921}\textsuperscript{\textpm 0.018}&
  \textbf{0.926}\textsuperscript{\textpm 0.017}\\
 &
  PPM &
  \textbf{0.976}\textsuperscript{\textpm 0.008} &
  {0.944}\textsuperscript{\textpm 0.011}*&
  {0.854}\textsuperscript{\textpm 0.035}*&
  {0.877}\textsuperscript{\textpm 0.021}*&
  {0.913}\textsuperscript{\textpm 0.017}*\\
 &
  TopicFit &
  {0.950}\textsuperscript{\textpm 0.013}* &
  {0.908}\textsuperscript{\textpm 0.014}* &
  {0.724}\textsuperscript{\textpm 0.022}* &
  {0.826}\textsuperscript{\textpm 0.028}* &
  {0.852}\textsuperscript{\textpm 0.017}* \\ 
  &
  O2D2 &
  0.904\textsuperscript{\textpm 0.023} &
  0.672\textsuperscript{\textpm 0.090} &
  0.458\textsuperscript{\textpm 0.085} &
  0.560\textsuperscript{\textpm 0.070} &
  0.648\textsuperscript{\textpm 0.059} \\ 
   &
  LUAR &
  0.964\textsuperscript{\textpm 0.008}&
  0.931\textsuperscript{\textpm 0.014}&
  0.775\textsuperscript{\textpm 0.035}&
  0.880\textsuperscript{\textpm 0.024}&
  0.887\textsuperscript{\textpm 0.017} \\ \hline
\multirow{5}{*}{Random} \rule{0pt}{2.5ex} & 
  CharNGram &
  0.966\textsuperscript{\textpm 0.008} &
  \textbf{0.962}\textsuperscript{\textpm 0.007} &
  0.863\textsuperscript{\textpm 0.027} &
  \textbf{0.918}\textsuperscript{\textpm 0.015} &
  \textbf{0.927}\textsuperscript{\textpm 0.011} \\
 &
  PPM &
  \textbf{0.977}\textsuperscript{\textpm 0.007} &
  0.955\textsuperscript{\textpm 0.009} &
  \textbf{0.879}\textsuperscript{\textpm 0.036} &
  0.893\textsuperscript{\textpm 0.022} &
  0.926\textsuperscript{\textpm 0.017} \\
 &
  TopicFit &
  0.961\textsuperscript{\textpm 0.011} &
  0.928\textsuperscript{\textpm 0.012} &
  0.754\textsuperscript{\textpm 0.033} &
  0.852\textsuperscript{\textpm 0.028} &
  0.874\textsuperscript{\textpm 0.019} \\
  &
  O2D2 &
  0.901\textsuperscript{\textpm 0.034} &
  0.712\textsuperscript{\textpm 0.132} &
  0.488\textsuperscript{\textpm 0.101} &
  0.583\textsuperscript{\textpm 0.088} &
  0.671\textsuperscript{\textpm 0.083} \\
  &
  LUAR &
  0.964\textsuperscript{\textpm 0.008} &
  0.935\textsuperscript{\textpm 0.020} &
  0.780\textsuperscript{\textpm 0.056} &
  0.876\textsuperscript{\textpm 0.040}	  &
  0.889\textsuperscript{\textpm 0.028} \\ \hline

\end{tabular}
\caption{Scores of AV models in HITS and randomly subsampled datasets (number of topics = 70), mean averaged across ten-fold validation splits. The best-performing models in each setup and metric are in \textbf{bold}. Asterisk (*) denotes significantly lower scores (p < 0.05 using unpaired t-tests) of HITS compared to Random.}
\label{mainresults}
\end{table*}
 
\subsection{Authorship verification methods}
In our experiments, we use various baselines from the PAN2021 competition \cite{kestemont2021overview} to assess the consistency and reliability of our HITS subsampled datasets compared to randomly distributed ones. We also experimented with two additional AV state-of-the-art models.

\paragraph{Character n-gram distance.} N-gram distance is a widely used baseline method in authorship verification. The character n-gram has also been used in previous studies \cite{stamatatos2013robustness, sapkota2014cross, stamatatos2017authorship} to achieve good performance in cross-topic scenarios. In our experiment, we use the implementation provided by the organizers of the PAN2021 competition \cite{kestemont2021overview}. We build the n-gram vocabulary set using our training data. At inference time, we compute the cosine similarity between each vector representation of each text in an input text pair. The similarity scores are then calibrated based on two thresholds (p1 and p2). This method performs a linear transformation as follows. Scores less than or equal to p1 are rescaled to the range [0, 0.49]. Scores between p1 and p2 are set to 0.5. Scores greater than or equal to p2 are rescaled to the range [0.51, 1]. p1 and p2 are hyperparameters we obtain using grid search on validation data held out from our training data.


\paragraph{Prediction by partial matching (PPM) \cite{teahan2003using}.} This model computes the cross-entropy of each text pair in the training data for each text pair (text1, text2). A compression model of text1 computes the cross-entropy of text2. The vice-versa is computed for text1. Afterward, the model computes the mean and absolute difference of the two cross-entropy values and predicts a probability score using logistic regression.

\paragraph{Topic-fit model.} We design a topic-fit model to assess the topic-shift effect of datasets, similar to the bias-only models used in studies of spurious correlations in natural language understanding tasks \cite{clark-etal-2019-dont, utama-etal-2020-mind, utama-etal-2020-towards, deutsch-etal-2021-towards}. However, our focus is that our topic-fit model is designed to fail when the topic in the test data changes from the training.

Our designed topic-fit model is a reversed version of the text distortion method \cite{stamatatos2017authorship}. The bias models are trained on input texts with top k most frequent features masked to obfuscate topic-independent words such as grammatical words. The non-masked words will likely be the content words, which should be more topic-dependent. For topic-fit models, we use the same implementation as the character n-gram baseline but with word unigrams instead of characters. The example of the masked input texts is illustrated in Table \ref{biasmodel}. We expect the resulting model to perform worse in cross-topic evaluation when topic-specific information is \emph{not} available in the test set.

\begin{table}[H]
\centering
\begin{tabular}{l|l}
\hline
version  & text                                                                                   \\ \hline
Original & \begin{tabular}[c]{@{}l@{}}The dogs and cats are \\ running in the garden\end{tabular} \\ \hline
Bias   & \begin{tabular}[c]{@{}l@{}}*** dogs *** cats *** \\ running ** *** garden\end{tabular} \\ \hline
\end{tabular}
\caption{An example of the masked input texts that we use to obtain a topic-fit model}
\label{biasmodel}
\end{table}



\paragraph{O2D2 \cite{boenninghoff2021o2d2}.} This approach uses CNN character embeddings and bidirectional LSTM with attention and modified contrastive loss to create text representation. They then use the representation with a combination of Bayes factor scoring, uncertainty adaptation, and out-of-distribution detector to predict the probability output. This framework achieved state-of-the-art on the PAN 2021 AV challenge. In our experiments, we train the O2D2 framework on our training data in each evaluation split using the authors' provided training scripts.

\paragraph{LUAR \cite{rivera-soto-etal-2021-learning}.} This model is based on a pre-trained SentenceBERT, fine-tuned using a Siamese network and supervised contrastive loss from aggregated sliding window text vectors. This model was not a part of the PAN2020/2021 challenge, but the authors used modified Fanfiction data from PAN2020/2021 in their study and achieved successful results. We train the LUAR framework on our training data in each evaluation split in our experiments using the authors' provided training scripts. Since the original author's setup does not directly predict the same-author probability of a text pair, we perform inference using the cosine similarity of the text pair from LUAR vectors, calibrated using the same method as the character n-gram baseline.

\paragraph{Evaluation.}
We follow the evaluation metrics used in the PAN 2020 and 2021 AV competitions \cite{Kestemont2020OverviewOT, kestemont2021overview}. Given an input pair of texts, models are expected to predict a score of between 0.0 and 1.0, indicating the probability of the text pair being written by the same author. Models are allowed to predict non-answers (scoring exactly at 0.5). The evaluation metrics used include the F1 score \cite{scikit-learn}, Area Under Receiver Operating Characteristic Curve (AUC) \cite{scikit-learn}, c@1  score \cite{penas-rodrigo-2011-simple, stamatatos2014overview}, and F0.5u score \cite{bevendorff-etal-2019-generalizing}, and Overall, the mean value of all other metrics. We use multiple metrics to allow comparison with the baseline results from the existing PAN2021 benchmark. Moreover, one might prioritize different properties of an AV system. For example, if one prioritizes precision over recall, F0.5u may be used over the F1 score. F0.5u and C@1 also reward the systems' ability to give non-answers in ambiguous cases, while the F1 score does not.

\section{Experimental Results}
\label{expresults}
This section illustrates the effectiveness of controlling topic similarity in mitigating topic leakage. We present a number of experiments comparing datasets sampled using the HITS method and randomly sampled datasets to see the difference between datasets with heterogeneous and randomly distributed topic sets. We present HITS results and randomly subsampled datasets, both with $m$ (number of topics) = 70. For Random datasets, we subsampled five datasets, each from a different random seed. The results for Random datasets in this section are reported with a mean average across five randomly subsampled datasets from different random seeds, each with a different topic set. On the other hand, since the HITS method is deterministic, there is only one subsampled dataset for the HITS dataset.

\textbf{Evaluation results.} First, we assess the performance of the baseline and state-of-the-art AV models on both setups. We present the mean average metrics across ten validation folds in Table \ref{mainresults}. 
We observe that most models have lower Overall scores in HITS than Random. 
This agrees with our hypothesis that controlling topic similarity helps reduce topic leakage, thus making the HITS test sets more challenging than Random datasets with more similar topics.
However, when we look at individual metrics, we found that the F1 scores of LUAR and CharNGram are higher in the HITS setup, while the c@1 and F0.5u metrics are lower on all models in the HITS setup. 
This different behavior might be caused by the fact that c@1 and F0.5u reward non-answers for difficult samples, but F1 doesn't.
Another observation is that the scores of topic-fit models are significantly lower on HITS than random in every metric, with the largest difference compared to other models.
This supports our hypothesis that the HITS dataset with reduced topic similarity allows the models relying on topic information to perform worse. 
However, even in random datasets where we expect topic leakage to happen, the topic-fit model does not gain enough advantage to outperform other models on this mean average results. 
Furthermore, we notice a low performance from O2D2 on both HITS and random datasets compared to the PAN2021 results, which we think is caused by the smaller training dataset from the subsampling process. This performance drop is also reported by \citet{brad-etal-2022-rethinking}, which used custom Fanfiction data splits in their study. 
Despite the limited data, O2D2 also demonstrates lower performance in HITS setup, like other models.
Our speculation for this result is that AV models gain an advantage from topic leakage in different degrees. 
Still, topic-fit models do not generalize as well as them, hence the lower scores. 

\textbf{Ranking stability.} We also assess the ranking stability of the HITS-sampled dataset in Table \ref{rankingstabmain}.
If the chance of topic leakage is decreased, then the models' rankings should be more consistent across validation splits.
We measure Spearman's rank correlation between model rankings on each pair of validation splits to assess the ranking stability. Then, we compute the mean average of the correlation values into a single figure for each metric. 
We found that the different topic sets in the dataset affect the ranking stability of models even when using k-fold cross-validation. 
For example, the Random dataset on seed 2 has 0.89 Spearman's rank correlation on average across metrics, but the correlation can be as low as 0.814 on seed 4. 
On the other hand, the HITS dataset with controlled topic similarity has a higher rank correlation than 4 out of 5 random datasets. It is higher than the average of all seeds of random datasets. 
Considering that real-world datasets without control for topic similarity might have high variance in ranking stability like the random datasets, it can be beneficial to use the HITS method (when applicable) to improve ranking stability.
As a side note, when comparing models evaluated on the HITS dataset, we think using metrics other than F1 is beneficial since its rank correlation is noticeably lower than other metrics.


\begin{table}[htbp]
\small
\centering
\resizebox{\columnwidth}{!}{%
\begin{tabular}{l|lllll|l}  Dataset 
                              &  AUC
                              &  c@1
                              &  F0.5u 
                              &  F1  
                              &  Overall 
                              & Average \\ \hline
HITS                 & \textbf{0.88} & \textbf{0.92} & \textbf{0.95}  & 0.88 & \textbf{0.94}    & \textbf{0.92}   \\
R\textsubscript{avg} & 0.78 & 0.88 & 0.89  & 0.87 & 0.87    & 0.86   \\
R\textsubscript{0}   & {0.79}* & {0.85}* & {0.86}*  & 0.86 & {0.91}*    & 0.85   \\
R\textsubscript{1}   & {0.81}* & {0.87}* & {0.92}*  & {0.82}* & {0.90}*    & 0.87   \\
R\textsubscript{2}   & {0.80}* & 0.91 & {0.92}*  & {0.92}* & {0.91}*    & 0.89   \\
R\textsubscript{3}   & {0.76}* & \textbf{0.92} & {0.91}*  & {\textbf{0.94}}* & {0.82}*    & 0.89   \\
R\textsubscript{4}   & {0.74}* & {0.86}* & {0.86}*  & {0.79}* & {0.88}*    & 0.81  
\end{tabular}%
}
\caption{Spearman's rank correlation of AV models compared between HITS and Random datasets, mean averaged across ten validation folds. The highest correlations are in \textbf{bold}. R\textsubscript{n} represents random datasets subsampled using seed $n$. R\textsubscript{avg} represents the mean average correlation between all random datasets. Asterisk (*) denotes a significantly lower correlation (p < 0.05 using unpaired t-tests) of each Random dataset compared to the HITS dataset.}
\label{rankingstabmain}
\end{table}

\textbf{Model ranking analysis.} While rank correlation might show the stability of the HITS dataset, we look further into the model rankings of each validation split. On average,  topic-fit models in HITS datasets have a lower average rank than in random. Other models have mixed rankings across Random datasets. Notably, CharNGram and PPM have mixed results on both HITS and Random datasets, while PPM has been reported as having higher performance on both PAN2020 and PAN2021 AV challenges. Unstable rankings between CharNGram and PPM illustrate that topic leakage in certain evaluation splits can change the performance ranking of models and might result in selecting models that might not perform the best on texts with unseen topics. Furthermore, we also notice O2D2 consistently being the lowest rank in all datasets and has no variation due to the small subsampled datasets' size.

\begin{table}[h]
\centering

\resizebox{\columnwidth}{!}{%
\begin{tabular}{l|lllll}
Dataset            
& CharNGram
& PPM 
& TopicFit 
& O2D2
& LUAR
 \\ \hline 
 \vspace{-0.4cm} & & & & & \\
HITS                 
& \textbf{1.2}\textsuperscript{\textpm 0.42}  
& 1.9\textsuperscript{\textpm 0.57}  
& 4.0\textsuperscript{\textpm 0.00}       
& 5.0\textsuperscript{\textpm 0.00}  
& 2.9\textsuperscript{\textpm 0.32}  
\\
R\textsubscript{avg} 
& \textbf{1.5}\textsuperscript{\textpm 0.58}
& 1.6\textsuperscript{\textpm 0.54}
& 3.8\textsuperscript{\textpm 0.43}     
& 5.0\textsuperscript{\textpm 0.00} 
& 3.1\textsuperscript{\textpm 0.64} 
\\
R\textsubscript{0}   
& 1.8\textsuperscript{\textpm 0.63}       
& \textbf{1.3}\textsuperscript{\textpm 0.48}
& 3.7\textsuperscript{\textpm 0.48}      
& 5.0\textsuperscript{\textpm 0.00} 
& 3.2\textsuperscript{\textpm 0.32}  
\\
R\textsubscript{1}   
& \textbf{1.3}\textsuperscript{\textpm 0.48}      
& 1.7\textsuperscript{\textpm 0.48}
& 3.7\textsuperscript{\textpm 0.48}    
& 5.0\textsuperscript{\textpm 0.00} 
& 3.3\textsuperscript{\textpm 0.48}  \\

R\textsubscript{2}   
& 1.6\textsuperscript{\textpm 0.52}       
& \textbf{1.4}\textsuperscript{\textpm 0.52} 
& 3.7\textsuperscript{\textpm 0.48}     
& 5.0\textsuperscript{\textpm 0.00}  
& 3.3\textsuperscript{\textpm 0.48}  
\\
R\textsubscript{3}   
& \textbf{1.5}\textsuperscript{\textpm 0.53}       
& \textbf{1.5}\textsuperscript{\textpm 0.53}
& 3.8\textsuperscript{\textpm 0.42}
& 5.0\textsuperscript{\textpm 0.00} 
& 3.2\textsuperscript{\textpm 0.42}  
\\
R\textsubscript{4}   
& \textbf{1.5}\textsuperscript{\textpm 0.71}      
& 1.9\textsuperscript{\textpm 0.57}
& 3.9\textsuperscript{\textpm 0.32}     
& 5.0\textsuperscript{\textpm 0.00} 
& 2.7\textsuperscript{\textpm 0.95}
\end{tabular}%
}%

\caption{Ranking of each model (lower is better) on each subsampled dataset, mean averaged across ten validation folds. The rankings are computed with the Overall metric. Top-ranked models in each dataset are in \textbf{bold}. R\textsubscript{n} represents random datasets subsampled using seed $n$. R\textsubscript{avg} represents the mean average between all five randomly subsampled datasets.}
\label{rankinganal}
\end{table}

\textbf{Subsampled topics examples.} We also question whether the subsampled topic set can be considered heterogeneous for human readers. Therefore, we look at the examples of topics used in the HITS dataset compared to random, presented in Table \ref{topicexamples}. We observe that the top similar train-test topics in Random datasets (more than 0.95) are much higher than the HITS ones (0.86-0.87). In addition, upon manual inspection, we found that 3 out of 5 most similar topics in Random datasets are what we consider closely related. For example, X-Men: Evolution and X-Men: The Movie fandoms are both from the X-Men franchise. Moreover, Star Wars and Star Wars: The Clone Wars are also from the same franchise. Furthermore,  Batman is also a subset of DC Superheroes. On the other hand, we do not find such patterns in the most similar topic pairs in HITS datasets other than all of them are based on Japanese fictional texts. 

\begin{table}[t]
\centering
\small
\begin{tabular}{lll}
\hline
\textbf{Train Topic}        & \textbf{Test Topic}                & \textbf{Sim}   \\ \hline
& \textbf{HITS} & \\ \hline
Girl Meets World   & One Tree Hill             & 0.872 \\
Durarara!! & Saiyuki                   & 0.862 \\
Naruto             & Tenchi Muyo               & 0.862 \\
Inuyasha           & Naruto                    & 0.862 \\
Gakuen Alice       & Naruto                    & 0.861 \\ \hline
& \textbf{Random} &  \\ \hline
 X-Men: Evolution   & X-Men: The Movie          & 0.971 \\
Star Wars          & Star Wars:  &  \\
& The Clone Wars & 0.961 \\
Final Fantasy I-VI & League of Legends         & 0.961 \\
Days of Our Lives  & General Hospital          & 0.961 \\
Batman             & DC Superheroes            & 0.951 \\ \hline
\end{tabular}%
\caption{Top 5 similar train-test topics from evaluation splits from HITS and random datasets. ``Sim'' denotes the cosine similarity between each train-test topic pair.}
\label{topicexamples}
\end{table}

\paragraph{Discussion.} The experimental results on HITS reveal the implications of mitigating topic leakage by controlling topic similarity. 
The efficacy of the HITS method in managing topic similarity adds reliability to model evaluation and selection processes.
The significantly lower scores of topic-fit and PPM models in HITS datasets (Table \ref{mainresults}) suggest that these models still rely on topic-specific information rather than generalizing to unseen topics. 
On the other hand, CharNGram outperforms other deep-learning-based models without a significant difference in performance between HITS and Random datasets. This finding suggests that the method is the best choice in datasets with the number of topics and authors comparable to our subsampled Fanfiction dataset, whether there is topic leakage or not.
In addition, Spearman's rank correlation in HITS datasets (Table \ref{rankingstabmain}) shows improved stability in model rankings compared to the Random datasets. This suggests the volatility of ranking models on datasets without control for topic similarity.
Moreover, the individual rankings are also influenced by randomness, as suggested in the model ranking analysis (Table \ref{rankinganal}). Even on the same number of topics, different randomly selected topics can result in CharNGram being the top performer in some cases and PPM in others.
Together, HITS datasets with controlled topic similarity can be valuable to ensure the reliable development of AV systems. As a result, we obtained improved accurate evaluation results and ranking stability compared to datasets without topic similarity control.
\section{Component Design Analysis}
\label{ablationsec}
In this section, we perform additional experiments to study the impact of varying the following components in the HITS method: 1. the number of topics, 2. whether to discard the unselected topics and 3. the choice of topic representation encoder.

\subsection{Effect of the number of topics}
\paragraph{Effect on dataset statistics.} First, we question the effect of parameter $m$ from our HITS method on the resulting subsampled dataset. As a result, we create multiple subsampled datasets with varying numbers of topics with both HITS methods and random sampling. 
The numbers of topics we studied are = [50, 60, 70, 80, 90, 100]. According to the data statistics in Table \ref{ffspstats}, an apparent effect is that the more topics in the subsampled corpus mean, the more documents available for both training and testing. There is also a larger number of authors as the topics increase. However, there is no noticeable difference in data statistics between HITS and random sampling.

\paragraph{Effect on topic similarity.} In addition, we want to know whether training and validation data (after the k-fold validation split) in the HITS dataset are more dissimilar than the randomly sampled datasets in different numbers of topics. To answer the question, we computed the mean and max topic cosine similarity of evaluation splits from subsampled datasets in different topic sizes. We also compare the topic similarity with randomly subsampled datasets. The topic similarity is illustrated in Table \ref{numtopicsim}. The general trend for topic similarity is that when the number of topics increases, the topic similarity increases. However, one notable difference is that the mean and max topic similarity in Random datasets are quite similar across all topic sizes. On the other hand, the topic similarity is much lower in HITS datasets at smaller numbers of topics and becomes closer to random datasets when the number of topics approaches 100. 

\begin{table}[h]
\small
\centering
\begin{tabular}{l|ll|ll}
\hline
Topics & Random   & HITS   & Random   & HITS  \\ \hline
                & \multicolumn{2}{l|}{Mean similarity} & \multicolumn{2}{l}{Max similarity} \\ \hline
50              & 0.836             & 0.766           & 0.920              & 0.837          \\
60              & 0.842             & 0.766           & 0.928             & 0.853          \\
70              & 0.841             & 0.775           & 0.922             & 0.862          \\
80              & 0.857             & 0.785           & 0.938             & 0.862          \\
90              & 0.862             & 0.795           & 0.950              & 0.868          \\
100             & 0.864             &   0.801           & 0.949             & 0.879  \\ \hline        
\end{tabular}%

\caption{Mean and max topic cosine similarity compared between random and HITS subsampled datasets after validation splits. The figures are the mean average of ten validation folds.}
\label{numtopicsim}
\end{table}

\begin{table}[h]
\centering
\resizebox{\columnwidth}{!}{%
\begin{tabular}{l|lllll|l} 
\hline
             & AUC  & C@1  & F0.5u & F1   & Overall & Avg. \\ \hline
50  & 0.88	 & 0.88	 & 0.88	 & 0.93	 & 0.91	    &  0.91	    \\
60  & 0.93 & 0.89 & 0.85  & 0.84 & 0.91    & 0.88    \\
70  & 0.88 & 0.92 & 0.95  & 0.88 & 0.94    & 0.92    \\
80  & 0.91 & 0.96 & 0.89  & 1.00 & 0.89    & 0.93    \\
90  & 0.91 & 0.93 & 0.91  & 1.00 & 0.89    & 0.93    \\
100 & 0.84   & 0.80   & 0.87    & 0.81   & 0.87      & 0.84     \\ \hline
\end{tabular}%
}
\caption{Spearman's rank correlation of AV models on datasets with numbers of topics using HITS subsampling. The Y-axis denotes the number of topics in the subsampled dataset. The X-axis denotes Spearman's rank correlation of ranks computed from each metric. ``Avg.'' denotes the mean Spearman's rank correlation across all metrics.}
\label{numtopicstab}
\end{table}

\paragraph{Effect on ranking stability.} Furthermore, we question how the changes in the number of topics affect evaluation regarding ranking stability. We computed Spearman's rank correlation similarly with the Section \ref{expresults}. The results are presented in Table \ref{numtopicstab}. One observation is that ranking stability seems to not correlate with topic similarity. The mean average Spearman's rank correlation across metrics starts at ~0.88 to 0.93 at 50 to 90 topics before falling off to 0.84 at 100 topics. With the exception of 60 topics, the rank correlations are similar.  We did not experiment with a smaller number of topics since if the number is too small, the result can be too random to be reliable due to the smaller dataset size. We also did not experiment with larger numbers of topics since when the number of topics is too large (in our case, 100), the topic similarity becomes close to that of the randomly sampled datasets.  We suggest tuning the number of topics or $m$ as a hyperparameter to obtain the best results, especially when applying the HITS method to other datasets.

\subsection{Topic sampling approach}
One may question whether it is reasonable to subsample a corpus into a topic-heterogeneous version since this method reduces the training and test data available for models. Therefore, we consider two different approaches.
\begin{itemize}
    \item \textbf{Cutting}. We select a set of $m$ topics from the entire topic set of the original dataset as described in Section \ref{proposedmethod}. This is the approach we use in our proposed method.
    \item \textbf{Grouping}. Instead of discarding the data in non-selected topics, we merge them with the nearest neighboring topic category to allow similar topics to prevent topic information leakage since the highly similar topics are either in training or test data together. 
\end{itemize}

\begin{table}[!h]
\small
\centering
\begin{tabular}{l|l|l}
\hline
Metric        & Cutting        & Grouping      \\ \hline 
AUC     & \textbf{0.884}	  & 0.851	\\
c@1     & \textbf{0.916}	& 0.904	\\
F\_0.5\_u  & \textbf{0.953} &	0.900	 \\
F1      & 	0.882 & \textbf{0.893}	 \\
Overall & \textbf{0.940}	& 0.900	 \\ \hline
Average & \textbf{0.915} &	0.890	\\ \hline
\end{tabular}
\caption{A comparison between Spearman's rank correlation (with p-value) across five random seeds between cutting and grouping approaches. The highest correlations are in \textbf{bold}. ``Average'' denotes the mean Spearman's rank correlation across all metrics.}
\label{ablationapproachrankcorr}
\end{table}

We compared Spearman's rank correlation between the cutting and grouping approach and presented the results in Table \ref {ablationapproachrankcorr}. This result shows that cutting yields a higher Spearman's rank correlation than grouping on c@1, F0.5u, F1, and Overall metrics. The ranking stability of the F1 metric is similar to grouping.  Our explanation for the lower-ranking stability of the grouping approach involves the dataset size. Since unselected topics are not discarded but merged with selected topics, more data is in the resulting subsampled dataset, thus more topic similarity and less stability. This finding also agrees with experiments on the number of topics, where Spearman's rank correlation degrades at a higher number of topics, which is also a larger dataset. One could also do a hybrid cutting-grouping approach, where only topics with similarity exceeding a certain threshold are merged. However, we did not experiment with such an approach due to the resources required for extensive threshold parameter tuning.

\subsection{Topic representation}
We consider the following vector representation mapping functions as candidates for creating topic representation for our sampling method: 
\begin{itemize}
    \item \textbf{Latent Dirichlet Allocation (LDA)} \cite{blei2003latent}. LDA is often used to perform topic modeling in an unsupervised manner. We hypothesize that we may be able to use the representation created by LDA to compare similarities between topics.
    \item \textbf{Non-Negative Matrix Factorization (NMF)}. It is another method commonly used for topic modeling. In a study by \citet{Kestemont2020OverviewOT, kestemont2021overview}, NMF has been used to test the models' correlation between text pairs' topic similarity and predicted results. 
    \item \textbf{SentenceBERT (sBERT)} \cite{reimers-gurevych-2019-sentence}. Studies have shown that fine-tuned pre-trained language models such as BERT \cite{devlin-etal-2019-bert} can create sentence representations that capture the semantic similarity between texts. 
\end{itemize}

\begin{table}[h]
\small
\centering
\begin{tabular}{l|l|l|l}
\hline
Metric   & sBERT   & LDA   & NMF \\ \hline
AUC    & \textbf{0.884}	& 0.436	& 0.667 \\
c@1     & \textbf{0.916} &	0.613 & 0.836  \\
F\_0.5\_u  &  \textbf{0.953} &	0.667 &	0.858  \\
F1      &  \textbf{0.882} &	0.809 &	0.702 \\ 
Overall &  \textbf{0.940} &	0.747 &	0.822 \\ \hline
Average & \textbf{0.915} &	0.654	&  0.777 \\ \hline
\end{tabular}%
\caption{A comparison between Spearman's rank correlation (with p-value) across five random seeds between LDA, NMF, and SentenceBERT representations. The highest correlations are in \textbf{bold}. ``Average'' denotes the mean Spearman's rank correlation across all metrics.}
\label{ablationvecrankcorr}
\end{table}

We compared Spearman's rank correlation between each candidate topic representation and presented the results in Table \ref{ablationvecrankcorr}. The experimental results reveal that HITS subsampling with SentenceBERT representation yields the most stable rankings on average. SentenceBERT outperforms other topic representations in all of the metrics. With these results, we select SentenceBERT as the topic representation for experiments in this paper. There is also an additional benefit: SentenceBERT is already pretrained and does not need to be trained specifically on the Fanfiction dataset.

\section{RAVEN Benchmark}
\label{benchmarksection}
We propose the \emph{Robust Authorship Verification bENchmark (RAVEN)} created with our HITS framework. The objective of our benchmark is to assess the robustness of authorship verification models by uncovering the topic bias, or their reliance on topic-specific features.  

\subsection{Benchmark description}
We use the same source dataset as our main experiments, the Fanfiction dataset from PAN2020/2021 competitions. Our benchmark consists of two sets of evaluation setups: Random and HITS. Each setup has ten evaluation splits comprising training and cross-topic test data. The data statistics of each version are described in Table \ref{ffspstats}. 

One could use the HITS-sampled data setup in the RAVEN benchmark the same way as a regular benchmark: select one of the evaluation splits, then train or fine-tune a system on the provided training data and evaluate on test data. However, we also propose another alternative evaluation method that might help uncover topic bias: the topic shortcut test.

\begin{table}[htbp]
\small
\centering
\begin{tabular}{l|llll}
\hline
Model  & Random & HITS & Avg.  & Diff      \\ \hline
Model1 & \textbf{0.80}    & 0.56 & 0.68  & 0.25    \\
Model2 & 0.75   & \textbf{0.72} & \textbf{0.73} & \textbf{0.02}  \\
Model3 & 0.76   & 0.70 & 0.72  & 0.06  \\ \hline
\end{tabular}
\caption{An example illustration is evaluating three different AV models using our RAVEN Benchmark. ``Random'' and ``HITS'' denote each model's average overall score (e.g. F1) across validation folds of random and HITS sampled datasets, respectively. ``Avg.'' denotes the mean average score  HITS and randomly sampled datasets (higher is better). ``Diff'' denotes the mean absolute difference in the scores of both setups (lower is better), which is intended to show the model's reliance on topic-specific information. The best models for each criterion are in \textbf{bold}}
\label{uncover}
\end{table}

\subsection{Topic shortcut test}
One might question how we can use the RAVEN benchmark to uncover the topic bias. The intuition behind using two evaluation setups is that models that rely on topic-specific features would perform worse in the heterogeneous split than the random one. This is similar to our experimental results in that topic-fit models reveal high score differences between HITS and randomly sampled datasets. To perform this test on a set of candidates AV systems, one follow the following steps: 

\begin{enumerate}
    \item Train and evaluate each system on our provided datasets, including each random seed of the HITS and randomly sampled datasets. 
    \item Aggregate the scores across random seeds into the mean average score.
    \item Compute the absolute difference between the mean average score of HITS sampled datasets and randomly sampled datasets.
\end{enumerate}

After these steps, we get results similar to our illustration in Table \ref{uncover}. We can use the mean absolute difference in score between these two sampling methods to uncover the topic bias in a model. Lastly, we can rank each score difference to select the most robust model against topic shift.

\section{Limitations}
It is important to address the limitations of our HITS evaluation method and the RAVEN benchmark.
First, the HITS method assumes a large number of topics and samples to still have sufficient data after removing some of the topics. One would also need to consider tuning the parameter $m$, which is the number of topics in the target subsampled dataset, to balance the trade-off between the dataset size and the degree of topic similarity.

Second, the HITS methods assume existing topic labels for each sample in the dataset. Our experiments use the topic label provided in the Fanfiction dataset. When applying the HITS method to other datasets without such labels, one needs to perform topic modeling methods to obtain the topics. However, the scope of our experiments does not cover the outcome of performing HITS on the automatically extracted topics.

Moreover, the score calibration used in some of the baselines in our experiments does not explicitly handle class imbalance, which might affect metrics such as the F1 score. When applying these baselines to other datasets, post-hoc calibration methods such as the one described by \mbox{\citet{guo2017calibration}} might be more suitable. We recommend exploring these calibration methods in future work to enhance the robustness of the score adjustments.

Furthermore, due to the subsampling process, the RAVEN benchmark is still limited in the number of topics, authors, and text samples. Therefore, this benchmark only simulates smaller real-world applications where a domain shift between training and inference, such as historical or literary texts, is expected. Future efforts can be made to improve the dataset size, which might better simulate other AV applications in large corpora.

\section{Conclusion}
In conclusion, we describe the topic leakage issue in the conventional cross-topic evaluation of authorship verification systems. We illustrate how topic leakage can cause misleading evaluation and unstable model rankings. 

To tackle these issues, we present HITS, an evaluation method that can create a dataset with heterogeneous topic sets from existing datasets. Our experimental results show that a heterogeneous topic set can help reduce topic information leakage, thus improving ranking stability in evaluating authorship verification models.

Furthermore, we present RAVEN, a benchmark created using the HITS method on the Fanfiction dataset. The benchmark is designed to uncover the degree of topic bias of authorship verification models to select the most robust one.  One can also use the HITS method on their datasets to create a similar benchmark.

\section{Reproducibility}
To allow the reproduction of our experiments and obtain the RAVEN benchmark, our source code for preprocessing, sampling, baseline authorship verification models, random seed, and other parameter settings is available at \href{https://github.com/jitkapat/hits\_authorship}{https://github.com/jitkapat/hits\_authorship}.


\bibliography{tacl2021}
\bibliographystyle{acl_natbib}

\appendix

\iftaclpubformat

\onecolumn

\appendix
\fi

\end{document}